\pdfoutput=1
\documentclass[11pt]{article}

\usepackage[final]{acl}
\usepackage{times}
\usepackage{latexsym}

\usepackage[T1]{fontenc}
\usepackage[utf8]{inputenc}
\usepackage{microtype}

\usepackage{inconsolata}

\usepackage{times}
\usepackage{latexsym}
\usepackage{graphicx}
\usepackage{subcaption}
\usepackage{epstopdf}
\usepackage{amsmath}
\usepackage{amssymb}
\usepackage{placeins}
\usepackage{authblk}
\usepackage{tabularx} 
\usepackage{multirow}
\usepackage{float}

\title{FreeCtrl: Constructing Control Centers with Feedforward Layers for Learning-Free Controllable Text Generation}

\author[1,3]{\textbf{Zijian Feng}}
\author[1,3]{\textbf{Hanzhang Zhou}}
\author[1,3]{\textbf{Zixiao Zhu}}
\author[2,3,\thanks{ \hspace{1mm} Corresponding author}]{\textbf{Kezhi Mao}}

\affil[1]{Institute of Catastrophe Risk Management, Interdisciplinary Graduate Programme, \authorcr Nanyang Technological University, Singapore}
\affil[2]{School of Electrical and Electronic Engineering, Nanyang Technological University, Singapore}
\affil[3]{Future Resilient Systems Programme, Singapore-ETH Centre, CREATE Campus, Singapore}
\affil[ ]{\texttt{\{feng0119, hanzhang001, zixiao001\}@e.ntu.edu.sg}, \texttt{ekzmao@ntu.edu.sg}}

\begin{document}
\maketitle
\begin{abstract}
% Controllable text generation (CTG) seeks to craft texts adhering to specific attributes, traditionally employing learning-based techniques such as training, fine-tuning, or prefix-tuning with attribute-specific datasets. These approaches, while effective, demand extensive computational and data resources. In contrast, some proposed learning-free alternatives circumvent learning but often yield inferior results, exemplifying the fundamental machine learning trade-off between computational expense and model efficacy. To overcome these limitations, we propose FreeCtrl, a learning-free method that dynamically modulates the weights of selected feedforward neural network (FFN) vectors to increase the likelihood of generating sentences with desired attribute-related keywords. Specifically, we first identify the key characteristics and challenges of using FFN layers for CTG and then introduce a structured workflow to build and adaptively activate control centers constructed by FFN vectors to regulate the language model outputs on desirable attributes. Extensive experiments on single- and multi-attribute control reveal that the proposed learning-free FreeCtrl outperforms other learning-free and learning-based methods, successfully resolving the dilemma between learning costs and model performance\footnote{Code is available at \url{https://github.com/zijian678/FreeCtrl}}.
Controllable text generation (CTG) seeks to craft texts adhering to specific attributes, traditionally employing learning-based techniques such as training, fine-tuning, or prefix-tuning with attribute-specific datasets. These approaches, while effective, demand extensive computational and data resources. In contrast, some proposed learning-free alternatives circumvent learning but often yield inferior results, exemplifying the fundamental machine learning trade-off between computational expense and model efficacy. To overcome these limitations, we propose FreeCtrl, a learning-free approach that dynamically adjusts the weights of selected feedforward neural network (FFN) vectors to steer the outputs of large language models (LLMs). FreeCtrl hinges on the principle that the weights of different FFN vectors influence the likelihood of different tokens appearing in the output. By identifying and adaptively adjusting the weights of attribute-related FFN vectors, FreeCtrl can control the output likelihood of attribute keywords in the generated content. Extensive experiments on single- and multi-attribute control reveal that the learning-free FreeCtrl outperforms other learning-free and learning-based methods, successfully resolving the dilemma between learning costs and model performance\footnote{Code is available at \url{https://github.com/zijian678/FreeCtrl}}.

\end{abstract}

\section{Introduction}

Controllable text generation (CTG) focuses on directing language models to produce diverse and fluent sentences that adhere to predefined single or multiple attributes such as topics and sentiment \citep{yang-etal-2023-tailor,gu-etal-2023-controllable,zhang2023survey, zhong-etal-2023-air}. Recent works on CTG can be roughly categorized into two groups based on their dependency on a learning process: learning-based methods and learning-free methods \citep{mireshghallah-etal-2022-mix}.

\begin{figure}  
\centering  
\includegraphics[scale=0.55]{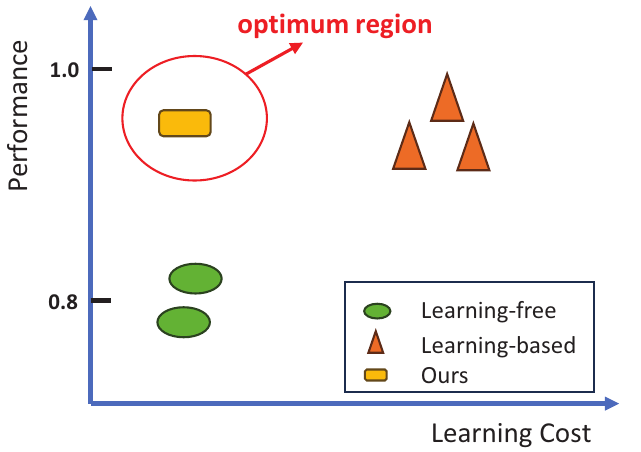}  
\caption{Trade-off between learning cost and performance for CTG. Learning-based methods excel in delivering superb results but demand significant training resources. Conversely, learning-free methods are more resource-efficient but tend to yield inferior performance. Numerical performance details are available in \S\ref{sec:exp}.}  
\label{fig_intro}  
\end{figure} 

Learning-based methods usually involve training \citep{yang-klein-2021-fudge, krause-etal-2021-gedi-generative, lin-riedl-2021-plug}, fine-tuning \citep{ficler-goldberg-2017-controlling, keskar2019ctrl, wang-etal-2021-mention}, or prefix-tuning  \citep{qian-etal-2022-controllable, zhang-song-2022-discup, gu-etal-2022-distributional,gu-etal-2023-controllable, yang-etal-2023-tailor, zhong-etal-2023-air} language models or discriminators based on attribute-specific data. Despite their effectiveness, these approaches come with high computational costs for training and a dependency on vast, attribute-specific datasets, posing challenges for deployment in environments with limited data or computational capacity. 

Only a few existing methods are learning-free, avoiding the need for training. For instance, K2T \citep{pascual-etal-2021-plug-play} employs attribute-focused keywords to influence token output probability during generation. Another method, Mix\&Match \citep{mireshghallah-etal-2022-mix}, integrates diverse black-box experts as a probabilistic energy model to steer large language model (LLM) outputs. These learning-free methods, despite bypassing the training process,  tend to fall short in performance compared to advanced learning-based approaches.

The analysis spotlights the classic dilemma in machine learning between the cost of learning and model performance, as shown in Figure \ref{fig_intro}. To overcome this obstacle, particularly in avoiding learning expenses for CTG while ensuring high performance, we propose \textbf{FreeCtrl}: Constructing Control Centers with Feedforward Layers for Learning-Free Controllable Text Generation. FreeCtrl's central idea is to manipulate FFN vectors\footnote{FFN vectors refer to the value vectors in the second weight matrix of the FFN layer. More details and definitions can be found in \S \ref{sec:theory}. } to regulate the outputs of LLMs, inspired by a recent finding that the tokens generated by LLMs can be attributed to the weights of vectors in FFN layers \citep{geva-etal-2022-transformer}.

%Specifically, recent studies \citep{geva-etal-2022-transformer} have shown that the tokens generated by LLMs can be attributed to the weights of vectors in FFN layers, suggesting that varying FFN vectors may influence the output probability of different tokens. Motivated by this, we investigate the potential to control FFN vectors to regulate LLM outputs.

%FreeCtrl's core concept involves pinpointing vectors in feedforward neural network (FFN) layers that regulate diverse attributes to create control centers for each. This method then controls LLM outputs on each attribute through a structured process of initialization, monitoring, adaptation, and filtering. Notably, this approach requires no training or attribute-specific data yet achieves or surpasses the efficacy of advanced learning-based models.

Specifically, the key principle is that increasing a single FFN vector's weight alters the output distribution, raising specific tokens' output probability. This strategy enables the targeted enhancement of certain FFN vector weights to raise attribute keywords' output probability, directing LLM generation towards preferred attributes. Our study first examines the possibility of this strategy by pinpointing three key characteristics of FFN vectors: 1) \textbf{Convergence}: Increasing the weight of an FFN vector can result in a stable and convergent output distribution in LLMs, thereby elevating the probabilities of specific tokens. 2) \textbf{Diversity}: Diverse FFN vectors can increase the output probabilities for most tokens in the LLM vocabulary, covering keywords relevant to general attributes in CTG; 3) \textbf{Prompt-Invariance}: the observed effects of convergence and diversity remain consistent across different input prompts. These characteristics suggest FFN vectors can enable stable, diverse controls for LLM outputs, directing sentence generation toward desired attributes.

However, we also identify a major limitation of FFN vectors: the \textbf{high-maintenance} challenge, where adjusting their weights for precise control proves difficult. Low weights lack the power to steer LLMs, while high weights compromise output diversity and fluency. 
To mitigate this, FreeCtrl initially sets up control centers using FFN vectors for various attributes, then navigates LLM output via a cycle of initialization, monitoring, adaptation, and filtering during the generation process. Continuous monitoring ensures that token generation is assessed at each step, allowing for adaptive weight adjustments of the control centers. Ultimately, a score-based filtering mechanism is employed to refine the outputs. Notably, this framework requires no training or attribute-specific data yet surpasses the efficacy of advanced learning-based models. Therefore, FreeCtrl addresses the cost-performance dilemma, situating it at the optimal upper-left corner in Figure \ref{fig_intro}, denoting learning-free but high performance. Our main contributions are summarized as follows:
\begin{enumerate}
    \item We conduct a systematic analysis of using FFN vectors for CTG in \S\ref{sec:ffnvec}, identifying three key characteristics: convergence, diversity, and prompt-invariance, alongside a notable challenge of high maintenance.
    \item We propose FreeCtrl in \S\ref{sec:method}, a learning-free approach that identifies and utilizes FFN vectors governing various attributes to establish control centers, thus enabling precise management of LLM outputs through initialization, monitoring, adaptation, and filtering.
    \item Comprehensive experiments in \S\ref{sec:exp} on both single and multi-attribute control demonstrate that  FreeCtrl, without incurring any learning costs, outperforms existing learning-free baselines and cutting-edge learning-based models.
\end{enumerate}

% Steward: Con\textbf{st}ructing Control C\textbf{e}nters with Feedfor\textbf{ward} Layers for Learning-Free Controllable Text Generation

\section{Related Work}

\textbf{Learning-based Methods} Initial research efforts concentrate on adapting language models into attribute-conditional language models, utilizing methods like fine-tuning \citep{ficler-goldberg-2017-controlling, keskar2019ctrl, wang-etal-2021-mention} and reinforcement learning \citep{ziegler2019fine, khalifa2020distributional, kim-etal-2023-critic}. Weighted decoding stands out as another key strategy in CTG. These methods are primarily learning-oriented, involving updates to the model's hidden states based on decoded logits \citep{dathathri2019plug} or training attribute discriminators to modify model output probabilities \citep{yang-klein-2021-fudge, krause-etal-2021-gedi-generative, lin-riedl-2021-plug}. Amidst the growth of large language models like GPTX \citep{radford2019language, openai_chatgpt} and LLaMA2 \citep{touvron2023llama}, recent techniques often preserve LLM parameters and utilize lightweight fine-tuning methods such as prefix-tuning \citep{li-liang-2021-prefix, lester-etal-2021-power} followed by decoding-time control \citep{qian-etal-2022-controllable, zhang-song-2022-discup, gu-etal-2022-distributional,gu-etal-2023-controllable, yang-etal-2023-tailor, zhong-etal-2023-air}. These methods generally necessitate a large volume of attribute-specific data and considerable computing resources for training either prefixes or discriminators.

\textbf{Learning-free Methods} The realm of learning-free approaches, which eschew any training process, is sparsely populated with research. One such example is K2T \citep{pascual-etal-2021-plug-play}, which shifts the output logit distribution by calculating the semantic similarity between vocabulary words and target attribute keywords. Notably, Mix\&Match \citep{mireshghallah-etal-2022-mix} stands as the pioneer in introducing a ``learning-free'' control framework. This innovation leverages external black-box scoring experts to evaluate the attributes of generated content, thereby regulating the model outputs. Compared to cutting-edge learning-based methods like PriorControl \citep{gu-etal-2023-controllable}, these approaches often fall short in performance.

\section{FFN Vectors for CTG}
\label{sec:ffnvec}

This section first details the theoretical basis for using FFN vectors to control LLM outputs, then examines their characteristics and challenges.

% Recent studies \citep{geva-etal-2022-transformer} have revealed that the output from each feedforward neural network (FFN) layer in Transformers contributes additively to the final prediction distribution. Building on this insight, our research explores the possibility of identifying distinct FFN vectors, each governing a different attribute. This could potentially allow for controlled text generation (CTG) through strategic manipulation of these FFN vectors.

\subsection{Theoretical Foundations}
\label{sec:theory}

In line with previous findings \citep{sukhbaatar2015end, sukhbaatar2019augmenting, geva-etal-2021-transformer, geva-etal-2022-transformer, fengunveiling}, the outputs from FFNs can be viewed as linear vector combinations:

\begin{align*}
\mathrm{FFN}^\ell(\mathbf{x}^\ell) &= f({W}_K^\ell \mathbf{x}^\ell) {W}_V^\ell \\ &= \sum_{i=1}^{d_m} f(\mathbf{x}^{\ell} \cdot \mathbf{k}_i)\mathbf{v}_i = \sum_{i=1}^{d_m} m_i^{\ell}\mathbf{v}_i
\end{align*}
where \( f \) is the activation function, \( {W}_K^\ell \) and \( {W}_V^\ell \) are the weight matrices, and \( \mathbf{x}^\ell \) is the input at layer \( \ell \). FFN then can be conceptualized as a neural key-value memory system, where the columns in \( W_K \) represent the keys and rows in \( W_V \) are the values. Given an input vector \( \mathbf{x}^{\ell} \), the keys generate the coefficients $\mathbf{m}^{\ell}=f\left(W_K^{\ell} \mathbf{x}^{\ell}\right) \in \mathbb{R}^{d_m} $, which assign weights to the values in \( W_V \).

% At layer \( \ell \), the FFN output is the product of two sequential linear transformations with an intermediate point-wise activation function, with bias terms omitted for simplicity:

% \[ \mathrm{FFN}^\ell(\mathbf{x}^\ell) = f({W}_K^\ell \mathbf{x}^\ell) {W}_V^\ell, \]
% where \( f \) is the activation function, \( \mathbf{W}_K^\ell \) and \( \mathbf{W}_V^\ell \) are the linear transformation weight matrices, and \( \mathbf{x}^\ell \) is the input vector at layer \( \ell \). FFN then can be conceptualized as a simulated neural key-value memory system, where the columns in \( W_K \) represent the keys and rows in \( W_V \) embody the values \citep{sukhbaatar2015end, sukhbaatar2019augmenting, geva-etal-2021-transformer, geva-etal-2022-transformer}. Given an input vector \( \mathbf{x}^{\ell} \), the keys generate the coefficients $\mathbf{m}^{\ell}=f\left(W_K^{\ell} \mathbf{x}^{\ell}\right) \in \mathbb{R}^{d_m} $ which serve to assign weights to the values in \( W_V \): 

%  \[ \mathrm{FFN}^{\ell}(\mathbf{x}^{\ell}) = \sum_{i=1}^{d_m} f(\mathbf{x}^{\ell} \cdot \mathbf{k}_i)\mathbf{v}_i = \sum_{i=1}^{d_m} m_i^{\ell}\mathbf{v}_i \]

In other words,  within each layer, the \textbf{value vectors} denoted as \( \mathbf{v}_i \) are extracted from the rows of the secondary weight matrix, \( W_V \). Taking  GPT2-medium \citep{radford2019language} as an example, \( W_V \) is dimensioned at \( 4096 \times 1024 \). This dimensionality implies the existence of 4096 value vectors, each extending to a 1024-dimensional space within each individual FFN layer. With 24 such layers incorporated within the GPT2-medium, the model encompasses \( 24 \times 4096 = 98,304 \) value vectors in total.

% To elucidate further, within each layer, the value vectors denoted as \( \mathbf{v} \) are extracted from the rows of the secondary weight matrix, \( W_V \). By examining the GPT2-medium model as a case study, \( W_V \) is dimensioned at \( 4096 \times 1024 \). This dimensionality implies the existence of 4096 value vectors, each extending to a 1024-dimensional space within each individual FFN layer. With 24 such layers incorporated within the GPT2-medium, the model encompasses \( 24 \times 4096 = 98,304 \) value vectors in total.

Prior research \citep{geva-etal-2022-transformer} has verified that the outputs generated by LLMs can be explained by examining the weights associated with the value vectors. The weights of various value vectors directly influence the probabilities of different token outputs. Building on this foundation, our study explores the identification of value vectors controlling different attributes in CTG and the feasibility of manipulating their weights to achieve attribute-specific control.

\begin{table*}[htbp]
\resizebox{\textwidth}{!}{
\begin{tabular}{ll}
\hline
\textbf{Attribute}  & \textbf{Keywords} \& \textbf{Positions}                                                                                     \\ \hline
POLITICS   & politics (20, 1651), government (22, 3127),  election (17, 1620), republic (0, 2991),   state (19, 84)    \\
SPORTS     & sports (14, 1078), champion (21, 4020), football (17, 573), game (23, 1928), coach (17, 1773)        \\ \hline
\end{tabular}
}
\caption{Politics and sports-related keywords along with their respective value vectors in GPT2. The position is denoted by $(a,b)$, where $a$ represents the layer number and $b$ is the position of the value vector within that layer.}
\label{tab_deiversity}
\end{table*}

\subsection{Characteristics of Value Vectors}
\label{subsec_character}

Effective control via value vectors depends on three core requirements: stable impact on output distributions,  the ability to manage a wide range of CTG attributes, and consistent behavior under different prompts. We highlight three key properties that affirm the value vectors' utility in achieving trustworthy CTG. Utilizing GPT2 as an example, we iteratively select a single value vector, denoted as $\mathbf{v}_i$, and then incrementally increase its weight, denoted as $u$. The resultant model output distribution, represented as $\mathbf{p}_i^u \in \mathbb{R}^{|\mathcal{V}|}$, is observed, where $\mathcal{V}$ signifies the GPT2's vocabulary and $|\mathcal{V}|$ is its size. 

\begin{figure}  
\centering  
\includegraphics[scale=0.40]{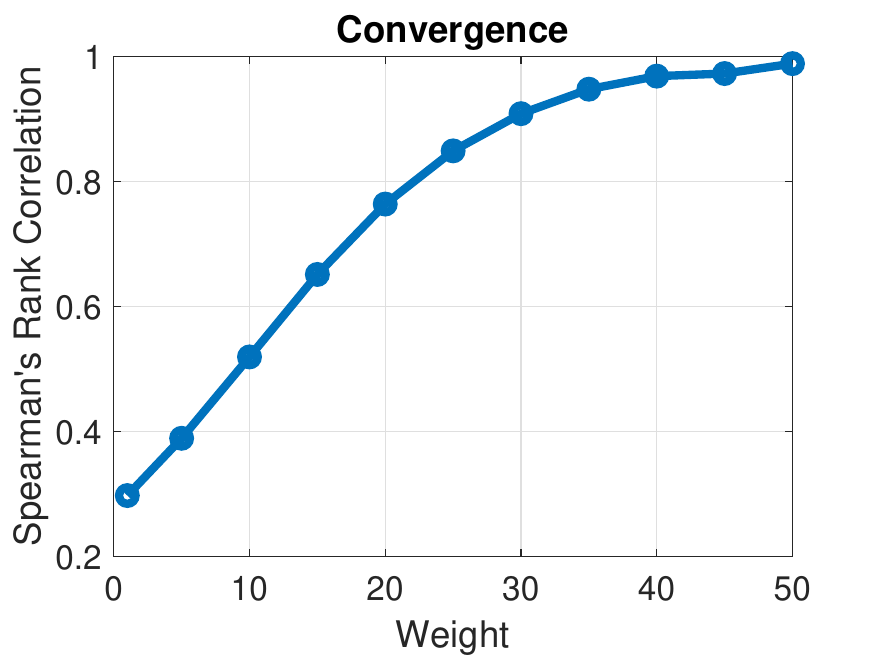}  
\caption{Convergence. As the value vector weight increases, its corresponding output distribution converges. }  
\label{fig_convergence}  
\end{figure}

\textbf{Convergence} While progressively increasing the weight $u$ from 1 to 50, the distribution influenced by each value vector progressively attains a state of stability and constancy, as shown in Figure \ref{fig_convergence}. Specifically, we treat the output distribution controlled by a weight of 50\footnote{A weight of 50 is considered exceptionally large according to \citet{geva-etal-2022-transformer}.} as the ground-truth $\mathbf{p}_i^g$ and calculate Spearman's rank correlation between distributions controlled by different weights and $\mathbf{p}_i^g$. The mean correlation across all 98,304 resultant distributions is reported in Figure \ref{fig_convergence}. Spearman's rank correlation is used because it directly compares token ranks and mitigates the impact of topk/p sampling and temperature variations. In summary, increasing the weights $u$ establishes stable token ranking and distribution patterns. Such convergence enables the discovery of patterns related to target attributes in CTG and facilitates stable control.

% \begin{figure}[htb]
%     \centering
%     \begin{subfigure}[b]{0.23\textwidth}
%         \centering
%         \includegraphics[width=\textwidth]{convergence.pdf}
%         \caption{Conventional}
%         \label{fig_intro1}
%     \end{subfigure}
%     \hfill
%     \begin{subfigure}[b]{0.23\textwidth}
%         \centering
%         \includegraphics[width=\textwidth]{diversity.pdf}
%         \caption{PromptExplainer}
%         \label{fig_intro2}
%     \end{subfigure}
%     \caption{Demonstration of conventional explanation methods and our proposed PromptExplainer. Conventional methods generally apply the linear operation to attentions and/or gradients to generate explanations, while PromtExplainer utilizes MLM head to disentangle token representations to explain language models.}
%     \label{fig_intro}
% \end{figure}

% \usepackage[normalem]{ulem}

\begin{figure}[h]  
\centering  
\includegraphics[scale=0.40]{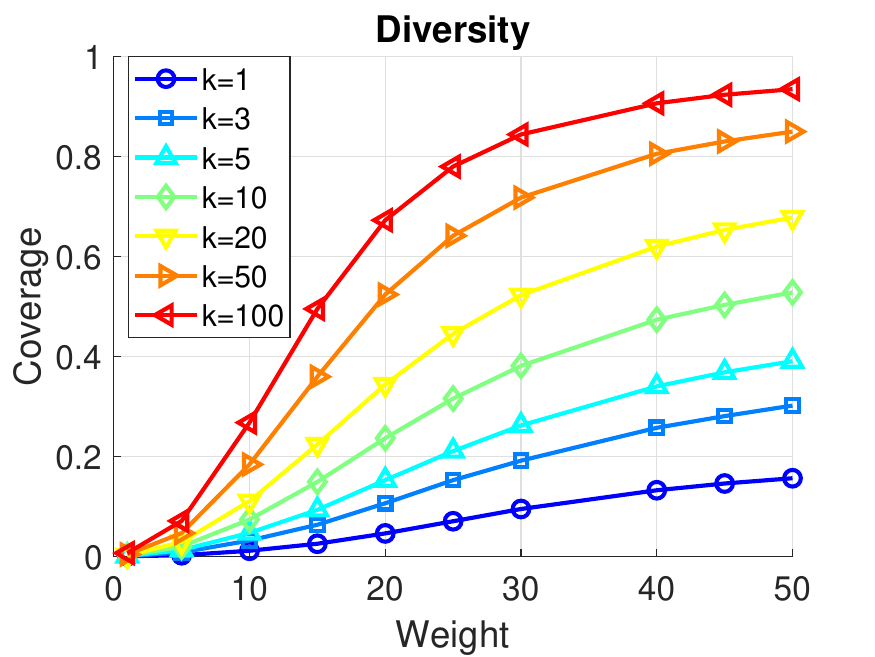}  
\caption{Diversity. The percentage of top-k tokens in the whole vocabulary grows with increasing weights. }  
\label{fig_diveristy}  
\end{figure}

\textbf{Diversity} The second question explores whether value vectors can sufficiently control a wide range of tokens representing various attributes in CTG. Our analysis is twofold: first, we assess the percentage of top-k controllable tokens in $\mathbf{p}_i^u$ over the whole covabulary; second, we identify specific value vectors that govern general attributes. Figure \ref{fig_diveristy} shows that with increasing weights, the top-1 tokens controlled by the 98,304 vectors account for up to 20\% of the GPT2 vocabulary, and this figure rises to over 80\% for top-50 tokens. This demonstrates the vectors' capacity to influence most tokens in the entire vocabulary. To provide further clarity, Table \ref{tab_deiversity} lists several value vectors alongside their corresponding attributes. More attributes and their associated vectors are detailed in Appendix \ref{sec:attributes}. These findings confirm the value vectors' ability to control a broad spectrum of attributes in CTG.

\textbf{Prompt-Invariance} To ensure effective control in LLMs, it is critical that value vectors maintain their properties across various input prompts. Our analysis involves feeding GPT2 with 35 different prompts, as provided by \citet{dathathri2019plug}. The results for various prompts mirror the earlier results, showcasing the characteristics of prompt-invariance.

\subsection{Limitation of High Maintenance}
\label{subsec:limitation}
While value vectors show promise for CTG, a significant challenge is their \textbf{high maintenance} due to the difficulty in setting optimal weights. Low weights fail to effectively direct LLMs towards desired attribute-specific tokens, whereas high weights can reduce output diversity and fluency. Furthermore, the ideal weights for various value vectors can be different. As illustrated in Appendix \ref{sec:app_highm}, a weight of 1 for politics-related vectors does not steer the model towards political content, but increasing the weight to 5 results in constrained and lower-quality outputs. Conversely, a weight of 5 is effective for sports-related attributes, producing relevant and high-quality generations.

\section{Methodology}
\label{sec:method}
\begin{figure*}  
\centering  
\includegraphics[scale=0.40]{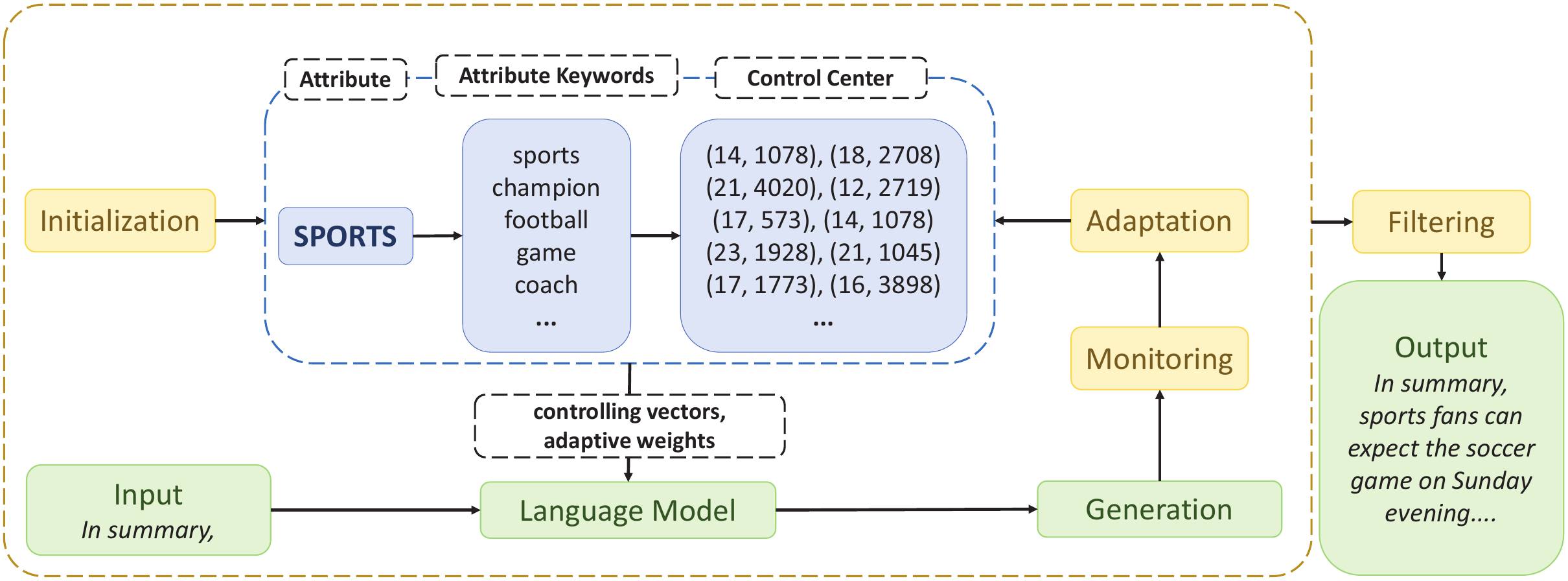}  
\caption{Overview of FreeCtrl. For the target attribute ``SPORTS'', FreeCtrl initially identifies related keywords and value vectors to establish a control center. Throughout the generation phase, it dynamically adjusts the control center's weights based on real-time output monitoring, ensuring adaptive feedback for subsequent token generation. Finally, a filter is applied to verify compliance with the required attribute. Notably, the position $(a,b)$ specifies the layer number $a$ and the value vector's position $b$ within that layer.}  
\label{fig_framework}  
\end{figure*}

To maximize the benefits and mitigate the limitations of FFN value vectors, we introduce FreeCtrl:  Constructing Control Centers with Feedforward Layers for Learning-Free Controllable Text Generation. FreeCtrl begins by gathering attribute keywords, subsequently constructing a control center for each attribute. It then guides the LLM to produce outputs relevant to the target attribute through a systematic pipeline of initialization, monitoring, adaptation, and filtering. The overall framework is illustrated in Figure \ref{fig_framework}.

\subsection{Attribute Keyword Collection}
For a given attribute $a_i$ in the attribute set $\mathbf{A} = \{a_1,\cdots,a_n \}$, we begin by gathering its associated keywords to promote diverse generations. Various external knowledge bases such as WordNet \citep{miller-1994-wordnet}, ConceptNet \citep{Speer_Chin_Havasi_2017}, RelatedWords\footnote{\url{https://relatedwords.org/}}, and ChatGPT \citep{openai_chatgpt} can be utilized for this purpose. To reduce the noises from these external sources, we apply a refinement function:

\begin{equation}
G(z) = r\left(z, a_i\right) \frac{|\mathbf{A}| - 1}{\sum_{a_j \in \mathbf{A}, a_j \neq a_i} r\left(z, a_j\right)}
\label{e_denoising}
\end{equation}
where $z$ represents the collected keyword for attribute $a_i$, and $r(\cdot)$ indicates the cosine similarity.

This function incorporates ideas from both TF-IDF \citep{sparck2004statistical} and KPT \citep{hu-etal-2022-knowledgeable}. It operates on the premise that a suitable attribute keyword should have a relevance score for its corresponding attribute that is higher than the average relevance score for other attributes. Consequently, keywords where $G(z) < 1.0$ are deemed less relevant and are filtered out for refinement. As this is not the primary focus and contribution of our work, we offer only a brief introduction here. For more detailed information, please refer to KPT \citep{hu-etal-2022-knowledgeable}.
%For an in-depth understanding, readers are referred to the KPT framework. As this concept is ancillary to our main focus, the discussion here is kept brief.

% !!!! PUT IN APPENDIX

\subsection{Control Center Construction}
Based on attribute-specific keywords, we can identify corresponding value vectors to direct the LLM toward outputs focused on these keywords and attributes. Building on \S \ref{subsec_character}, we iteratively assign a weight of 50 to each vector and examine the output distribution to locate dominant value vectors that control the attribute keywords.

% Formally,  let $\mathbf{P}_{u_{max}} \in \mathbb{R}^{N \times |\mathcal{V}|}$ denote the softmaxed output distribution over the vocabulary $\mathcal{V}$ regulated by value vectors of weight $u_{max}=50$, the control effect for the keyword $z$ can be obtained by $\mathbf{P}_{u_{max}}[:,d_{\mathcal{V}}(z)] \in \mathbb{R}^N$, where $d_\mathcal{V}(z)$ is the index of the token $z$ in the vocabulary and $N$ is the number of value vectors. Then we can build the control center by choosing the value vectors with the top-$k$ controlled probabilities:

Formally, let $\mathbf{P}_{u_{max}} \in \mathbb{R}^{N \times |\mathcal{V}|}$ represent the softmaxed output distribution across the vocabulary $\mathcal{V}$, modulated by all the $N$ value vectors with a set weight of $u_{max}=50$. The control effect for a specific keyword $z$ is captured by $\mathbf{P}_{u_{max}}[:,d_{\mathcal{V}}(z)] \in \mathbb{R}^N$, where $d_\mathcal{V}(z)$ denotes the index of token $z$ in the vocabulary, and $N$ signifies the total number of value vectors. To pinpoint vectors controlling the keyword $z$, we choose value vectors with top-$k$ probabilities:

\begin{equation}
\mathbf{c}_z = d_{vec}\{\max_k (\mathbf{P}_{u_{max}}[:,d_{\mathcal{V}}(z)])\}
\label{e_location}
\end{equation}
where $d_{vec}(\cdot)$ retrieves the index of value vectors in $N$. 
For instance, the positions (21,4020) and (12,2719) in Figure \ref{fig_framework} of value vectors correspond to the top-2 output probabilities for the attribute keyword ``champion''.

Finally, the control center for an attribute $a_i$ is established by aggregating the value vectors of the keywords related to the attribute $a_i$:

\begin{equation}
\mathbf{C}_{a_i} = \bigcup{\mathbf{c}_z, z \in \mathcal{Z}(a_i)}
\label{e_collection}
\end{equation}
where $\mathcal{Z}(a_i)$ denotes the set of keywords for the attribute $a_i$.

\subsection{Single-Attribute Control}

To precisely control LLMs through control centers, we adopt a structured process that includes initialization, monitoring, adaptation, and filtering. We constantly monitor the LLM's generation and then adaptively adjust control parameters to steer output toward the desired attribute. A final filtering step verifies the output's compliance with the specified attribute.

\textbf{Initialization} For a specified attribute $a_i$, we first locate value vectors to construct the control center $\mathbf{C}_{a_i}$. 

\textbf{Monitoring} At this stage, each token produced by the LLM in response to a prompt is evaluated for its relevance to the attribute $a_i$. We construct current output $\mathbf{s}_t^{a_i}$ by integrating the input prompt with tokens generated by the LLM at timestamp $t$ for attribute $a_i$. The embedding for token $s_i$ in $\mathbf{s}_t^{a_i}$ is derived as $E[s_i] \in \mathbb{R}^{d_e}$, where $E[\cdot]$ is the LLM's embedding matrix with dimension $d_e$. The attribute embedding $E[\mathcal{Z}(a_i)]$ can be obtained in a similar way. The correlation between current output $\mathbf{s}_t^{a_i}$ and target attribute $a_i$ can be calculated as:

\begin{equation}
\rho_t^{a_i} = \frac{1}{l_t} \sum_{j=1}^{l_t} \max  \{r( E[s_j], E[\mathcal{Z}(a_i)]\}
\label{e_corr}
\end{equation}
where $l_t$ denotes the length of the current sentence and the correlation score $\rho_t^{a_i}$ ranges from 0 to 1. This equation initially computes the maximum cosine similarity between each token in the current output and all target attribute keywords, subsequently calculating the mean value of these correlation scores. 

To enhance the correlation with the target attribute while minimizing it with other attributes, we utilize Eq.\ref{e_sentence_score}, which resembles Eq.\ref{e_denoising} and is designed to calculate the score of the current output. By continuously tracking the sentence score $\mu_t^{a_i}$, we are able to quickly modify the weights of $\mathbf{C}_{a_i}$, ensuring a coherent and seamless output that aligns with the targeted attribute.

\begin{equation}
\mu_t^{a_i} =\rho_t^{a_i} \frac{|\mathbf{A}| - 1}{\sum_{a_j \in \mathbf{A}, a_j \neq a_i} \rho_t^{a_j}}
\label{e_sentence_score}
\end{equation}

\textbf{Adaptation} Utilizing $\mu_t^{a_i}$, we can dynamically adjust the weights for timestamp $t+1$, facilitating smooth control and generation. To clarify, the model is required to generate a token at each timestamp. The primary reason for this adaptation is the high maintenance of value vector weights, as discussed in \S \ref{subsec:limitation}. The weight for timestamp $t+1$ is determined as Eq.\ref{e_weight}. Here, $\mu_{\omega}$ denotes a preset hyperparameter of the sentence score, $\lambda$ is a scaling parameter, and $\mu_{s_{l_t}}^{a_i}$ is the last token score obtained by Eqs.\ref{e_corr} and \ref{e_sentence_score}. We regard $\widehat{\mu}_t^{a_i} = \max(\mu_t^{a_i},\mu_{s_{l_t}}^{a_i})$ as final score to ensure the fluency.

% that establishes the threshold for sentence scores as calculated by Eq. \ref{e_sentence_score}. The parameter $\omega_{\text{cutoff}}$ refers to the minimum weight threshold, such as 0 or 0.1, ensuring that the LLM does not produce outputs that are adverse to the target attribute.

% \begin{equation}
%  \widehat{\mu}_t^{a_i} = \max(\mu_t^{a_i},\mu_{s_t}^{a_i}) >0
% \label{e_true_sen_score}
% \end{equation}

\begin{equation}
\omega_{t+1}^{a_i} = \begin{cases}
\frac{\lambda}{1+\exp[{-(\mu_{\omega}-\widehat{\mu}_t^{a_i})\cdot l_t}]} &  \mu_{\omega}-\widehat{\mu}_t^{a_i} >0 \\
0  & \text{otherwise}
\end{cases}
\label{e_weight}
\end{equation}

To clarify, a value of $\mu_{\omega}-\widehat{\mu}_t^{a_i} >0 $ indicates that the score of the current sentence or the last-generated token is below the predefined threshold, and as a result, the weight should be determined by the difference between $\mu_{\omega}$ and $\widehat{\mu}_t^{a_i}$. Additionally, the sentence length $l_t$ implies that weights at the outset of generation will be higher than those assigned later. This is grounded in the rationale that initially higher weights are necessary to guide the generation towards the desired direction. Once this direction is established, the LLM tends to continue generating tokens along this path, allowing for reduced weights in later stages to maintain fluency. Conversely, when $\mu_{\omega}-\widehat{\mu}_t^{a_i} < 0 $, it signifies that the sentence or the last-generated token at timestamp $t$ has an adequate score and aligns with the target attribute, eliminating the need for a higher weight. Setting the weight to 0 is a deliberate strategy to prevent the LLM from generating outputs associated with contrary attributes.
 
\textbf{Filtering} Through continuous monitoring and adaptation, the LLM can be steered to generate outputs focused on target attributes. However, some generations might not meet the $\mu_{\omega}$ threshold throughout the generation process, despite maintaining high weights. To filter out such non-compliant generations, a final screening is conducted using Eq.\ref{e_filtering}. Only those sentences that achieve scores in accordance with Eq.\ref{e_filtering} are considered valid outputs.

\begin{equation}
\mu_T^{a_i} > \mu_{\omega}
\label{e_filtering}
\end{equation}
where $T$ represents the final timestamp in the generation process, with a token produced at each timestamp.

% c_i^{forward} = \begin{cases}
% r_i & \text{if } i =1 \\
% r_i - r_{i-1} & \text{if } i>1
% \end{cases}

\begin{table*}[h!]
\resizebox{\textwidth}{!}{
\begin{tabular}{lccccccccccc}
\hline
\multicolumn{1}{c|}{\multirow{2}{*}{\textbf{Methods}}} & \multicolumn{3}{c|}{\textbf{Sentiment}$\uparrow$ (\%)}                                                    & \multicolumn{5}{c|}{\textbf{Topic}$\uparrow$ (\%)}                                                                                                        & \multicolumn{1}{c|}{\multirow{2}{*}{\begin{tabular}[c]{@{}c@{}}\textbf{Detox.}\\ $\uparrow$(\%)\end{tabular}}} & \multicolumn{1}{c|}{\multirow{2}{*}{\begin{tabular}[c]{@{}c@{}}\textbf{PPL}\\ $\downarrow$\end{tabular}}} & \multicolumn{1}{c}{\multirow{2}{*}{\begin{tabular}[c]{@{}c@{}}\textbf{Dist.-1/2/3}\\ $\uparrow$\end{tabular}}} \\ \cline{2-9}
\multicolumn{1}{c|}{}                         & \multicolumn{1}{l|}{\textbf{Avg.}} & \multicolumn{1}{l|}{\textbf{Neg.}} & \multicolumn{1}{l|}{\textbf{Pos.}} & \multicolumn{1}{l|}{\textbf{Avg.}} & \multicolumn{1}{l|}{\textbf{P.}} & \multicolumn{1}{l|}{\textbf{S.}} & \multicolumn{1}{l|}{\textbf{B.}} & \multicolumn{1}{l|}{\textbf{T.}} & \multicolumn{1}{c|}{}                                                                       & \multicolumn{1}{c|}{}                                                                    & \multicolumn{1}{c}{}                                                                            \\ \hline
\hline
\multicolumn{12}{l}{\textit{Learning-based   Methods}}                                                                                                                                                                                                                                                                                                                                                                                                                                                                                                                    \\
\textbf{PPLM}                                          & 80.0                      & 97.2                      & 62.7                      & 70.6                      & 74.9                    & 46.5                    & 62.4                    & 98.6                    & 93.2                                                                                        & 63.2                                                                                     & 31.1/70.9/85.9                                                                                  \\
\textbf{GeDi}                                          & 88.4                      & 96.6                      & 80.2                      & 90.8                      & 84.3                    & 92.6                    & 87.1                    & 99.2                    & 95.4                                                                                        & 134.1                                                                                    & 47.5/88.9/93.0                                                                                  \\
\textbf{Contra. Prefix}                                & 89.5                      & 88.4                      & 90.6                      & 86.7                      & 74.5                    & 85.3                    & 93.5                    & 93.6                    & 93.8                                                                                        & 37.7                                                                                     & 17.3/47.0/71.1                                                                                  \\
\textbf{Discrete}                                      & 92.5                      & 99.1                      & 85.9                      & 90.4                      & 84.5                    & 95.0                    & 84.6                    & 97.5                    & 90.1                                                                                        & 46.2                                                                                     & 36.9/76.3/87.0                                                                                  \\
\textbf{PriorControl}                                  & 97.1                      & \textbf{99.9}                      & 94.3                      & 95.9                      & \textbf{95.5}           & \textbf{99.3}           & 90.2                    & 98.7                    & 90.7                                                                                        & 54.3                                                                                     & 29.1/70.1/86.9                                                                                  \\ \hline \hline
\multicolumn{12}{l}{\textit{Learning-free   Methods}}    
\\  
\textbf{Mix\&Match}                                    & 82.8                      & 99.2                      & 63.3                      & 75.6                      & 79.5                    & 57.4                    & 69.6                    & 99.3                    & 96.9                                                                                        & 65.2                                                                                     & 31.5/74.8/88.8                                                                                  \\
\textbf{FreeCtrl (Ours)}                               & \textbf{97.7}             & \textbf{99.9}             & \textbf{95.4}             & \textbf{96.5}             & 93.7                    & 96.1                    & \textbf{96.5}           & \textbf{99.6}           & \textbf{97.3}                                                                               & 27.2                                                                                     & 20.2/61.3/84.1                                                                                  \\ \hline
\end{tabular}
}
\caption{Automatic results on single-attribute control. Results are reported for the attributes of \textbf{Pos}itive, \textbf{Neg}ative, \textbf{P}olitics, \textbf{S}ports, \textbf{B}usiness, \textbf{T}echnology, and \textbf{Detox}ification, in addition to the computed \textbf{Av}era\textbf{g}e score. Fluency is measured using perplexity (\textbf{PPL}), and diversity (\textbf{Dist-1/2/3}) is evaluated by distinct uni-, bi-, and tri-grams.}
\label{tab_single_control}
\end{table*}

\subsection{Multi-Attribute Control}

A significant strength of our method is its seamless adaptability to controlling multiple attributes. Specifically, in the case of multi-attributes $\{a_1,\cdots, a_{M} \}$, we initially gather the respective control centers for these attributes based on Eqs.\ref{e_location} and \ref{e_collection}. We then calculate sentence scores and weights for all \(M\) attributes by Eqs.\ref{e_corr}- \ref{e_weight}. At each timestamp, we select the control center with the highest weight for control. Assuming the control center $\mathbf{C}_{a_m}$ for the $m$-th attribute $a_m$ has the maximum weight $\omega_{t+1}^{a_m}$, then the weight for \(\mathbf{C}_{a_m}\) is set to \(\omega_{t+1}^{a_m}\), while weights for control centers of all other attributes are set to 0. Through this process, different control centers are activated at different timestamps, ultimately yielding an output that integrates multiple attributes after undergoing a filtering process as Eq.\ref{e_filtering}.

\section{Experiments}
\label{sec:exp}

\begin{table*}[h]
\resizebox{\textwidth}{!}{
\begin{tabular}{lcccccc}
\hline
\textbf{Methods}         & \textbf{Average $\uparrow$ (\%)} & \textbf{Sentiment $\uparrow$ (\%)} & \textbf{Topic $\uparrow$ (\%)} & \textbf{Detoxification $\uparrow$ (\%)} & \textbf{PPL.} $\downarrow$ & \textbf{Dist.} $\uparrow$ \\ \hline \hline
\multicolumn{7}{l}{\textit{Learning-based   Methods}}                                                                                                                       \\ 
\textbf{PPLM}            & 71.0 ± 21.4             & 64.7 ± 24.8                 & 63.5 ± 22.7           & 84.9 ± 6.5                       & 62.6          & 62            \\
\textbf{GeDi}            & 81.4 ± 14.7             & 76.1 ± 17.2                 & 73.8 ± 11.3           & 94.2 ± 1.9                       & 116.6         & 75.1          \\
\textbf{Contra. Prefix}  & 81.3 ± 16.5             & 74.4 ± 19.6                 & 76.9 ± 16.7           & 92.7 ± 3.5                       & 31.9          & 43.3          \\
\textbf{Discrete}        & 87.4 ± 10.9             & 86.7 ± 10.5                 & 84.8 ± 14.2           & 90.7 ± 7.4                       & 28.4          & 49.5          \\
\textbf{PriorControl}    & 89.9 ± 8.7              & 88.0 ± 10.6                 & 87.4 ± 8.5            & 94.3 ± 3.2                       & 34.7          & 55.5          \\ \hline \hline
\multicolumn{7}{l}{\textit{Learning-free   Methods}}                                                                                                                        \\ 
\textbf{Mix\&Match}      & 79.7 ± 21.8             & 73.5 ± 25.9                 & 69.9 ± 21.1           & \textbf{95.8 ± 1.9 }                      & 63.0          & 61.8          \\
\textbf{FreeCtrl (Ours)} & \textbf{93.4 ± 6.9}     & \textbf{95.7 ± 8.4}         & \textbf{89.7 ± 5.8}   & 94.7 ± 2.2              & 25.7          & 53.4          \\ \hline
\end{tabular}
}
\caption{Automatic evaluation results on multi-attribute control. 
The overall and individual average scores for sentiments, topics, and detoxification are reported. ± denotes the standard deviation, which reflects the stability of models among different attribute combinations}
\label{e_multi_attri}
\end{table*}

\subsection{Experimental Setups}

To align with established methods for CTG and ensure fair comparisons, our experimental setups rigorously follow Discrete \citep{gu-etal-2022-distributional} and PriorControl\citep{gu-etal-2023-controllable}.

\textbf{Tasks} Our analysis includes three CTG tasks: topic, sentiment, and detoxification, under both single- and multi-attribute control scenarios. Following PPLM \citep{dathathri2019plug}, we utilize 35 neutral prompts. The GPT2-medium \citep{radford2019language} is employed to generate sentence completions. For single-attribute control, GPT2 produces 5 completions for each attribute across all prompts, culminating in a total of 35 prompts $\times$ (2+4+1) attribute scenarios $\times$ 5 completions=1225 sentences. In the multi-attribute control, the model generates a total of 1,400 sentences, calculated as 35 prompts $\times$ (2$\times$4$\times$1) attribute combinations $\times$ 5 completions.

\textbf{Implementation Details} Implementation details and hyperparameter settings for our methods and comparative baselines are detailed in Appendix \ref{app:implement}.

% \textbf{Implementation Details} Learning-based methods typically require extensive attribute-specific datasets. In line with prior studies, we provide them with the AGNews \citep{zhang2015character}, IMDB \citep{maas2011learning}, and Jigsaw Toxic datasets \footnote{\url{https://www.kaggle.com/c/jigsaw-toxic-comment-classification-challenge/}} for topics, sentiments, and detoxification, respectively. Our approach is learning-free and obviates the need for training datasets. Following KPT \citep{hu-etal-2022-knowledgeable}, we gather and refine topic-attribute keywords using RelatedWords and source sentiment-related keywords for positive and negative attributes from the AFINN \citep{nielsen2011new} sentiment lexicon. Our method constructs a control center using positive versus toxic keywords from \citet{gehman-etal-2020-realtoxicityprompts} for single-attribute detoxification and filters out toxic words from negative keywords to enable the generation of non-toxic, negative content for multi-attribute control. In this way, each attribute contains between 200 to 300 keywords. Our proposed FreeCtrl features three hyperparameters: the number of value vectors $k$ for each attribute keyword in Eq.\ref{e_location}, the sentence threshold $\mu_{\omega}$, and the scaling factor $\lambda$ in Eq.\ref{e_weight}. The specific values applied in our experiments are detailed in Appendix \ref{app:hyper_para}.

\textbf{Baselines} We compare our FreeCtrl with both learning-based and learning-free methods. The learning-based approaches include (1) \textbf{PPLM} \citep{dathathri2019plug}, which leverages classifiers' gradients as bias indicators to guide the model's outputs; (2) \textbf{GeDi} \citep{krause-etal-2021-gedi-generative}, which steers the decoding stage using compact conditional generative models; (3) \textbf{Contrastive Prefix} \cite{qian-etal-2022-controllable}, incorporating contrastive learning into the prefix strategies to control the LLM generations; (4) \textbf{Discrete} \citep{gu-etal-2022-distributional}, employing discrete sampling to map the distribution of attributes within latent space to guide the LLM output; and (5) \textbf{PriorControl} \citep{gu-etal-2023-controllable}, which transfers complex distributions as simple Gausian distributions by using normalizing flow. For learning-free baselines, we compare with the advanced \textbf{Mix\&Match} \citep{mireshghallah-etal-2022-mix}, which uses external scoring experts to assess generated content attributes.

\textbf{Evaluation} We conduct both automatic and human evaluation. For \textbf{automatic evaluation}, we leverage classifiers from prior research \citep{gu-etal-2022-distributional, gu-etal-2023-controllable} to assess topic relevance and sentiment accuracy. Additionally, we utilize the Google Perspective API \footnote{\url{https://www.perspectiveapi.com}} to evaluate the effectiveness of detoxification.  We also report the generation fluency using the mean perplexity and diversity calculated by the mean number of unique n-grams \citep{li-etal-2016-diversity}. In \textbf{human evaluation}, three annotators evaluate each output based on text quality and the relevance of the specified attribute. These elements are scored on a 1 to 5 scale, where higher scores signify superior performance.

\subsection{Single-Attribute Control}

% \usepackage{multirow}

Table \ref{tab_single_control} presents the automatic evaluation results on single-attribute control. When compared to the advanced learning-free approach Mix\&Match, our proposed method, FreeCtrl, demonstrates superior performance across all attributes, with average improvements of 14.9\% in sentiment control, 20.9\% in topic control, and 0.4\% in detoxification. These results distinctly showcase FreeCtrl's significant advancement over current state-of-the-art (SOTA) learning-free techniques in CTG. Compared to learning-based approaches, FreeCtrl demonstrates competitive or superior performance against the SOTA PriorControl. Specifically, FreeCtrl achieves an average improvement of 0.6\% over PriorControl in both sentiment and topic control domains and significantly outpaces PriorControl by a notable margin of 6.6\% in detoxification. The results from human evaluation, as shown in Table \ref{tab_human_eval}, further reveal a similar trend as automatic evaluations. It is noteworthy that FreeCtrl operates without the need for a learning/training phase or training data, yet it still secures the best results. This underlines FreeCtrl's potential in addressing the challenge of optimizing the balance between cost and performance, as depicted in Figure \ref{fig_intro}.

\begin{table}[]
\resizebox{0.45\textwidth}{!}{
\begin{tabular}{lccc}
\hline
\textbf{Method}          & \textbf{Quality}$\uparrow$ & \textbf{Attribute}$\uparrow$ & \textbf{Avg.}$\uparrow$ \\ \hline
\multicolumn{4}{l}{\textit{Single-Attribute Control}}                            \\ \hline
\textbf{Mix\&Match}      & 3.2              & 3.4                & 3.3           \\
\textbf{PriorControl}    & \textbf{4.2}     & 4.3                & \textbf{4.3}  \\
\textbf{FreeCtrl (Ours)} & 4.1              & \textbf{4.5}       & \textbf{4.3}  \\ \hline
\multicolumn{4}{l}{\textit{Multi-Attribute Control}}                             \\ \hline
\textbf{Mix\&Match}      & 3.0              & 3.1                & 3.1           \\
\textbf{PriorControl}    & \textbf{3.9}              & 4.1                & 4.0           \\
\textbf{FreeCtrl (Ours)} & 3.8              & \textbf{4.3}       & \textbf{4.1}  \\ \hline
\end{tabular}
}
\caption{Human evaluation results. Quality and Attribute are assessed on a 1 to 5 scale, focusing on text quality and relevance to the specified attribute. The inter-annotator agreement is 0.33 based on Fleiss’ Kappa. }
\label{tab_human_eval}
\end{table}

\subsection{Multi-Attribute Control}

Table \ref{e_multi_attri} details the results of multi-attribute control evaluations, where FreeCtrl markedly outperforms both the learning-based SOTA PriorControl and the learning-free SOTA Mix\&Match by significant margins. Specifically, FreeCtrl exceeds Mix\&Match's performance by 22.2\% and PriorControl's by 7.7\% in sentiment control, and by 19.8\% and 2.3\% in topic control, respectively. Furthermore, FreeCtrl enhances the overall average score by 13.7\% over Mix\&Match and by 3.5\% over PriorControl. 
The human evaluation results presented in Table \ref{tab_human_eval} further highlight the superior performance of our method. These findings underscore FreeCtrl's efficiency in CTG, demonstrating its capability to excel without relying on a training set or undergoing a learning process.

\begin{table*}[]
\centering
% \small
\resizebox{0.75\textwidth}{!}{

\begin{tabular}{lccccccc}
\hline
\textbf{Method}        & \textbf{P. ↑} & \textbf{S. ↑} & \textbf{B. ↑} & \textbf{T. ↑} & \textbf{Avg. ↑}       & \textbf{PPL ↓}        & \textbf{Dist ↑}         \\
\hline
FreeCtrl      & 93.7 & 96.1 & 96.5 & 99.6 & 96.5         & 28.9         & 20.2/61.3/84.1 \\
Ran. Mon      & 86.1 & 83.2 & 84.2 & 95.5 & 88.0 (-8.5)  & 31.1 (+2.2)  & 16.2/51.7/75.5 \\
w/o Ada (0.5) & 74.9 & 79.2 & 77.8 & 92.1 & 81.0 (-15.5) & 15.7 (-13.2) & 25.6/59.2/83.7 \\
w/o Ada (1.5) & 88.5 & 90.3 & 96.4 & 98.4 & 93.4 (-3.1)  & 39.9 (+11.0) & 15.2/46.2/70.4 \\
w/o Fil       & 89.7 & 88.6 & 91.7 & 98.3 & 92.1 (-4.4)  & 26.4 (-2.5)  & 19.9/59.4/81.2\\
\hline
\end{tabular}
}
\caption{The ablation study on topic control using different components of FreeCtrl.}
\label{e_ablation}
\end{table*}

\subsection{Ablation Study}
To elucidate the impact of each FreeCtrl component, we conduct an ablation study focusing on topic control as follows:

\begin{itemize}
    \item Random Monitoring (\textbf{Ran. Mon}): We replace the monitoring score in Eq. \ref{e_sentence_score} with a random score generator, which produces scores uniformly distributed between 0 and 2.
    \item Without Adaptation (\textbf{w/o Ada}): In this variant, we remove the adaptation component entirely and apply static weights (1.5 and 0.5) to examine how the system performs without the ability to adjust control weights dynamically based on monitoring feedback.
    \item Without Filtering (\textbf{w/o Fil}): This setup tests FreeCtrl's performance without the filtering process.
\end{itemize}

Table \ref{e_ablation} summarizes the results, demonstrating that replacing or removing monitoring, adaptation, or filtering components leads to performance drops. Specifically, using a constant weight of 0.5 reduces performance by 15.5\% but increases diversity due to reduced control. Conversely, a constant weight of 1.5 significantly lowers both fluency and diversity, indicating that excessive control in LLMs is detrimental. These outcomes highlight the critical role of adaptive control in FreeCtrl for balancing performance, fluency, and diversity in text generation.

% \begin{table*}[]
% \begin{tabular}{llllllll}
% Method        & P. ↑ & S. ↑ & B. ↑ & T. ↑ & Avg. ↑       & PPL ↓        & Dist ↑         \\
% FreeCtrl      & 93.7 & 96.1 & 96.5 & 99.6 & 96.5         & 28.9         & 20.2/61.3/84.1 \\
% Ran. Mon      & 86.1 & 83.2 & 84.2 & 95.5 & 88.0 (-8.5)  & 31.1 (+2.2)  & 16.2/51.7/75.5 \\
% w/o Ada (0.5) & 74.9 & 79.2 & 77.8 & 92.1 & 81.0 (-15.5) & 15.7 (-13.2) & 25.6/59.2/83.7 \\
% w/o Ada (1.5) & 88.5 & 90.3 & 96.4 & 98.4 & 93.4 (-3.1)  & 39.9 (+11.0) & 15.2/46.2/70.4 \\
% w/o Fil       & 89.7 & 88.6 & 91.7 & 98.3 & 92.1 (-4.4)  & 26.4 (-2.5)  & 19.9/59.4/81.2
% \end{tabular}
% \end{table*}

\subsection{Diversity Analysis}

The experimental results presented in Tables \ref{tab_single_control} and \ref{e_multi_attri} reveal a slight decrease in the diversity of generated content. To mitigate this, we propose increasing the temperature settings of the LLMs during generation. Tables \ref{tab_temp_single} and \ref{tab_temp_mul} display the outcomes of applying our FreeCtrl method with an increased temperature setting, compared against other prominent baselines for both single-attribute and multi-attribute control tasks. The results demonstrate that increasing the temperature enables FreeCtrl to achieve the highest diversity scores while also maintaining superior control accuracy and fluency, as evidenced by Perplexity (PPL) scores.

\begin{table}[h]
\centering
\resizebox{0.49\textwidth}{!}{
\begin{tabular}{lcccccc}
\hline
\textbf{Method} & \textbf{Sentiment} & \textbf{Topic} & \textbf{Detox.} & \textbf{PPL} & \textbf{Dist.} \\
\hline
Discrete            & 92.5 & 90.4 & 90.1 & 46.2 & 66.7 \\
PriorControl        & 97.1 & 95.9 & 90.7 & 54.3 & 62.0 \\
Max\&Match          & 82.8 & 75.6 & 96.9 & 65.2 & 65.0 \\
FreeCtrl & 97.4 & 96.1 & 97.1 & 39.3 & 71.2 \\
\hline
\end{tabular}
}
\caption{Results of increased temperature on single-attribute control.}
\label{tab_temp_single}
\end{table}

\begin{table}[h]
\centering
\resizebox{0.49\textwidth}{!}{

\begin{tabular}{lccccc}
\hline
\textbf{Method} & \textbf{Sentiment} & \textbf{Topic} & \textbf{Detox.} & \textbf{PPL} & \textbf{Dist.} \\
\hline
Discrete            & 86.7 & 84.8 & 90.7 & 28.4 & 49.5 \\
PriorControl        & 88.0 & 87.4 & 94.3 & 34.7 & 55.5 \\
Max\&Match          & 73.5 & 69.9 & 95.8 & 63.0 & 61.8 \\
FreeCtrl  & 95.8 & 88.9 & 95.0 & 34.6 & 68.1 \\
\hline
\end{tabular}
}
\caption{Results of increased temperature on multi-attribute control.}
\label{tab_temp_mul}
\end{table}

\subsection{Further Analysis}
% In this section, we present a comprehensive analysis of our proposed FreeCtrl. Initially, we assess its scalability across larger language models, specifically LLaMA2-7B and LLaMA2-13B \citep{touvron2023llama}. Subsequently, we examine the impact of the three hyperparameters on control effectiveness. Lastly, we offer case studies to illustrate the generated outputs and the dynamic adjustment of controlling weights throughout the process.

% In this section, we analyze FreeCtrl, assessing its scalability on LLaMA2-7B and LLaMA2-13B \citep{touvron2023llama}, examining hyperparameter control effects, and presenting case studies on outputs and adaptive weight adjustments.

Further analysis is provided as follows:
\begin{itemize}
    \item Hyperparameter analysis: We examine three hyperparameters in FreeCtrl for adjusting the control strength in Appendix \ref{app:hyper}.
    \item Case study: For a visual illustration of control effects, output examples along with their corresponding control weights are presented in Appendix \ref{app:case_study}.
    \item Inference speed: Given that monitoring, adaptation, and filtering could add additional time costs, we assess FreeCtrl's inference speed and compare it with other methods in Appendix \ref{app:inference}.
    \item Scalability: To verify its scalability, we extend FreeCtrl to a larger language model, LLaMA2-7B \citep{touvron2023llama}, with detailed results presented in Appendix \ref{app:scalability}.
\end{itemize}

\section{Conclusions}
In this paper, we introduce FreeCtrl, a learning-free approach for controllable text generation (CTG). FreeCtrl employs FFN value vectors to establish control centers tailored to each attribute, enabling dynamic control via a structured process of initialization, monitoring, adaptation, and filtering. Comprehensive experiments demonstrate that FreeCtrl markedly outperforms both learning-based and learning-free methods. %effectively resolving the trade-off between learning costs and model performance.

\section{Limitations}
Our FreeCtrl approach effectively navigates the trade-off between learning expenses and model efficacy. We believe its control mechanism could be further streamlined while maintaining satisfactory outcomes. Additionally, delving deeper into the dynamics of value vectors, including their interactions, can enrich our comprehension and enhance CTG design strategies. These areas offer promising directions for future research.

\section*{Acknowledgements}
We express our sincere gratitude to the reviewers for their insightful and constructive feedback. We would like to acknowledge that this project is an outcome of the Future Resilient Systems initiative at the Singapore-ETH Centre (SEC). Additionally, we extend our thanks to the National Research Foundation, Prime Minister’s Office, Singapore, for their invaluable support through the Campus for Research Excellence and Technological Enterprise (CREATE) programme.

\bibliography{acl_latex}

\appendix

\section{Attributes and Corresponding Value Vectors in GPT2}
\label{sec:attributes}

Table \ref{tab_ak} details the general attribute keywords in CTG along with their associated value vectors.

\begin{table*}[]
\resizebox{\textwidth}{!}{
\begin{tabular}{ll}
\hline
\textbf{Attribute}  & \textbf{Keywords} \& \textbf{Positions}                                                                                     \\ \hline
POLITICS   & politics (20, 1651), government (22, 3127),  election (17, 1620), republic (0, 2991),   state (19, 84)    \\
SPORTS     & sports (14, 1078), champion (21, 4020), football (17, 573), game (23, 1928), coach (17, 1773)       \\
BUSINESS   & business (21, 1631), commerce (16, 2225), trade (17, 3938), market (22,   876), finance (22, 2709)        \\
TECHNOLOGY & technology (0, 3260), engineering (0, 3780), science (13, 3160), internet   (15, 547), robotics (0, 3260) \\
POSITIVE   & admire (10, 459), great (23, 318), wonderful (12, 3475), good (20, 841),   happy (20, 2959)               \\
NEGATIVE   & worse (17, 3792), bad (19, 3834), abuse (23, 2534), corrupt (0, 2890),   fake (21, 1027)                  \\
FOOD       & food (21, 3922), rice (14, 423), meat (15, 3011), milk (19, 2113), salt   (13, 1992)                      \\
AMERICAN   & America (19, 684), us (12, 3116), Trump (16, 558), bush (22, 819),   American (23, 1417)                  \\
ASIAN      & Asia (2, 1409), Japan (18, 1794), Korea (7, 2880), Singapore (18, 1794),   China (19, 3818)               \\
COMPUTER   & laptop (19, 741), hardware (16, 1933), cpu (4, 283), processor (18,   3717), disk (18, 2619)              \\
MILITARY   & military (14, 2816), war (6, 989), army (23, 3142), navy (23, 1396),   soldier (11, 469)                  \\
LEGAL      & legal (18, 1137), court (19, 999), justice (18, 4022), legislation (15,   596), rule (21, 634)            \\
RELIGION   & religion (18, 3564), faith (21, 3294), god (8, 1710), bless (20, 691),   church (14, 3094)                \\ \hline
\end{tabular}
}
\caption{Commonly-used attribute keywords and their corresponding positions in GPT2. The position is denoted by $(a,b)$, where $a$ represents the layer number and $b$ signifies the position of the value vector within that layer.}
\label{tab_ak}
\end{table*}

\section{Examples of High Maintenance}
\label{sec:app_highm}

\begin{table*}[]
\resizebox{\textwidth}{!}{
\begin{tabular}{lll}
\hline
\textbf{Attribute} & \textbf{Weight} & \textbf{Output}                                                                                          \\ \hline
politics  & 1.0      & This essay discusses there is sufficient evidence to support the   conclusion that there is ... \\
politics  & 3.0      & This essay discusses political philosophy, including how philosophy can   aid us as ...         \\
politics  & 5.0      & This essay discusses a state of state mind is a state of state...                               \\
sports    & 5.0      & This essay discusses soccer in America. It's about the beautiful  games that we watch...        \\ \hline
\end{tabular}
}
\caption{GPT2 outputs controlled by value vectors of different weights. The input prompt is ``This essay discusses''.}
\label{tab_highm}
\end{table*}

We begin by identifying value vectors linked to particular attributes and then vary their weights to assess the impact on LLM outputs, as summarized in Table \ref{tab_highm}. A weight of 1 for politics-related vectors is insufficient to direct the model's focus towards political themes. Elevating the weight to 5 leads to diminished output diversity and quality. In contrast, applying a weight of 5 to sports-related vectors successfully generates relevant and high-quality content. These results verify the high maintenance of value vectors.

\section{Implementation Details}
\label{app:implement}

Learning-based methods typically require extensive attribute-specific datasets. In line with prior studies, we provide them with the AGNews \citep{zhang2015character}, IMDB \citep{maas2011learning}, and Jigsaw Toxic datasets \footnote{\url{https://www.kaggle.com/c/jigsaw-toxic-comment-classification-challenge/}} for topics, sentiments, and detoxification, respectively. Our approach is learning-free and obviates the need for training datasets. Following KPT \citep{hu-etal-2022-knowledgeable}, we gather and refine topic-attribute keywords using RelatedWords\footnote{\url{https://relatedwords.org/}} and source sentiment-related keywords for positive and negative attributes from the AFINN \citep{nielsen2011new} sentiment lexicon. Our method constructs a control center using positive versus toxic keywords from \citet{gehman-etal-2020-realtoxicityprompts} for single-attribute detoxification and filters out toxic words from negative keywords to enable the generation of non-toxic, negative content for multi-attribute control. In this way, each attribute contains 200 to 300 keywords. Our proposed FreeCtrl features three hyperparameters: the number of value vectors $k$ for each attribute keyword in Eq.\ref{e_location}, the sentence threshold $\mu_{\omega}$, and the scaling factor $\lambda$ in Eq.\ref{e_weight}. 
Hyperparameter configurations for single- and multi-attribute control experiments are detailed in Table \ref{tab_hyper-para}.
% the number of value vectors $k$ for each attribute keyword in Eq.\ref{e_location}, the sentence threshold $\mu_{\omega}$, and the scaling factor $\lambda$ in Eq.\ref{e_weight}. The specific values applied in our experiments are detailed in Appendix C.

\begin{table}[]

\begin{tabular}{llll}
\hline
Hyperparameter  & $k$    & $\mu_{\omega}$     &  $\lambda$    \\ \hline
\multicolumn{4}{l}{Single -Attribute} \\ \hline
Topic           & 30   & 1.15  & 1.5  \\
Sentiment       & 30   & 1.15  & 0.3  \\
Detoxification           & 30   & 1.15  & 0.3  \\ \hline
\multicolumn{4}{l}{Multi-Attribute}   \\ \hline
Topic           & 200  & 1.1   & 0.5  \\
Sentiment       & 200  & 1.1   & 0.5  \\
Detoxification           & 200  & 1.1   & 0.5  \\ \hline
\end{tabular}
\caption{Hyperparameter setting for single- and multi-attribute control tasks.}
\label{tab_hyper-para}
\end{table}

% \section{Scalability on LLaMA2}

\section{Hyperparameter Analysis}
\label{app:hyper}

Our proposed FreeCtrl has three hyperparameters: the number of value vectors $k$ for each attribute keyword in Eq.\ref{e_location}, the sentence threshold $\mu_{\omega}$, and the scaling factor $\lambda$ in Eq.\ref{e_weight}. 

\begin{figure}  
\centering  
\includegraphics[scale=0.40]{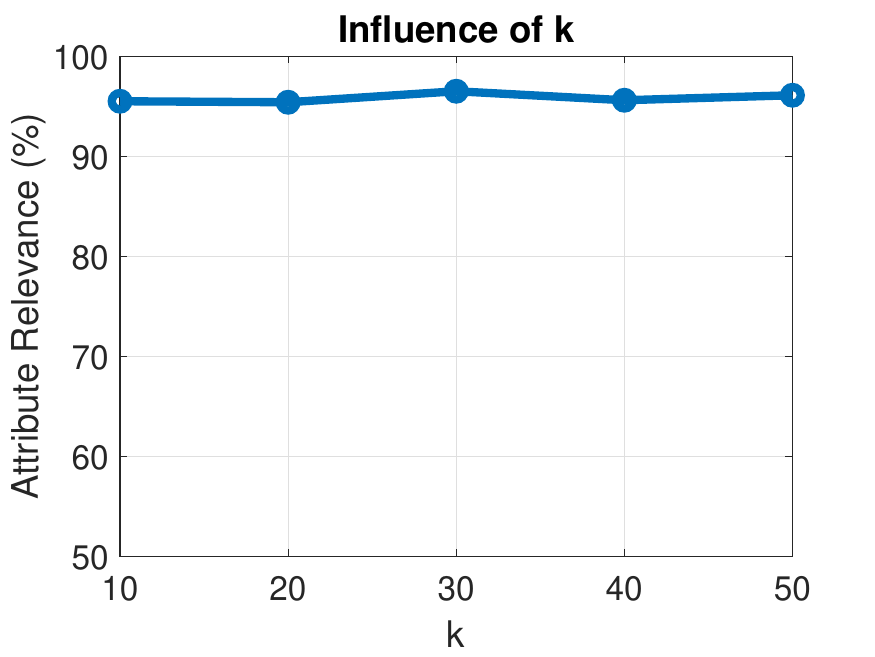}  
\caption{Influence of $k$ on topic control. }  
\label{fig_k_influence}  
\end{figure}

The hyperparameter $k$ represents the number of value vectors used to regulate a single attribute keyword. Empirical evidence suggests that $k=30$ for single-attribute control and $k=200$ for multi-attribute control yield satisfactory results. The necessity for a greater number of value vectors in multi-attribute control arises from the increased complexity and heightened competition among attributes. Figure \ref{fig_k_influence} outlines the average impact of varying $k$ on topic control. Observations reveal that when $k$ ranges from 10 to 50, the control effects fluctuate slightly between 95.4\% and 96.5\%, illustrating the robustness of FreeCtrl.

\begin{figure}  
\centering  
\includegraphics[scale=0.40]{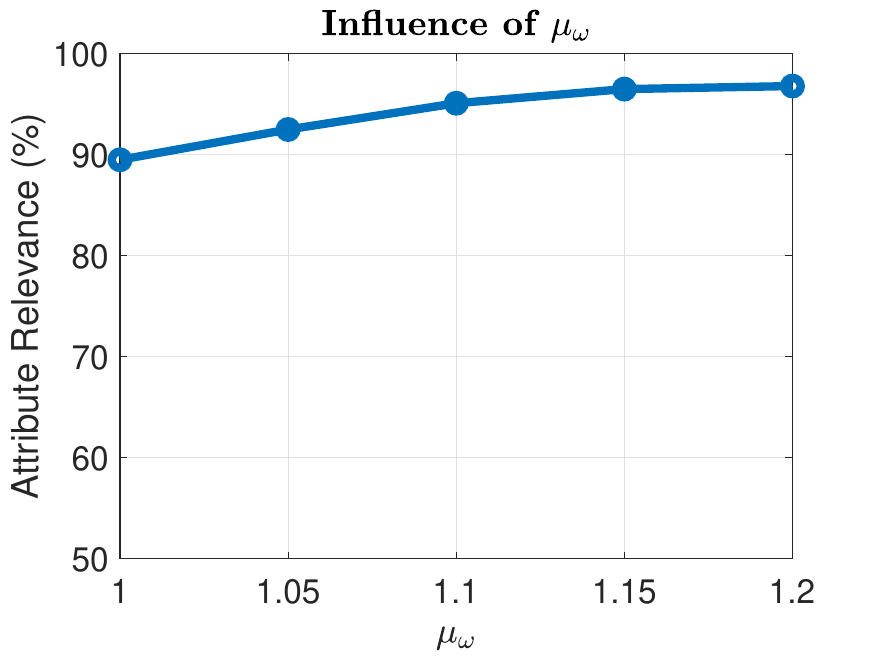}  
\caption{Influence of $\mu_{\omega}$ on topic control. }  
\label{fig_miu_influence}  
\end{figure}

The second hyperparameter, $\mu_{\omega}$, defines the sentence score threshold for control, affecting the final output collection. Figure \ref{fig_miu_influence} shows the effect of altering $\mu_{\omega}$ between 1.0 and 1.2 on topic control effectiveness. As $\mu_{\omega}$ increases from 1.0 to 1.1, there is a gradual improvement in performance. Adjusting $\mu_{\omega}$ further, from 1.1 to 1.2, results in stable and satisfactory performance, ranging between 95\% and 97\%.

\begin{figure}  
\centering  
\includegraphics[scale=0.40]{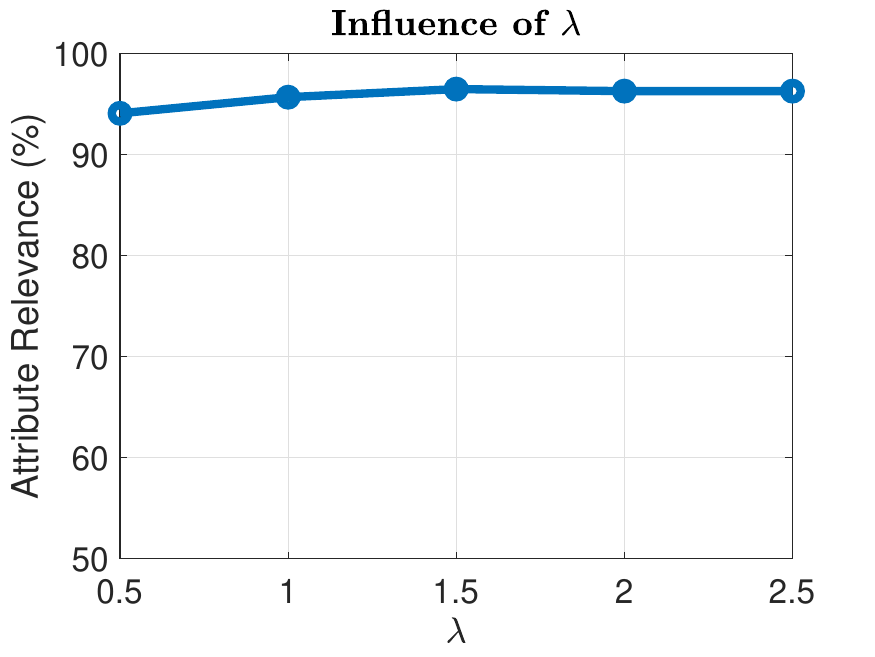}  
\caption{Influence of $\lambda$ on topic control. }  
\label{fig_lambda_influence}  
\end{figure}

The third hyperparameter, $\lambda$, acts as a scaling factor for the control weight. Figure \ref{fig_lambda_influence} indicates that setting $\lambda$ to 0.5 yields a 94.1\% effectiveness in topic control. As $\lambda$ increases from 1.0 to 2.5, performance improves, ranging between 95.7\% and 96.5\%. This suggests that a higher $\lambda$ enhances the control effect. Consistently achieving over 95.7\% effectiveness with $\lambda$ above 1.0 underscores FreeCtrl's efficacy and robustness.

\section{Case Study}
\label{app:case_study}

\begin{figure*}  
\centering  
\includegraphics[scale=0.50]{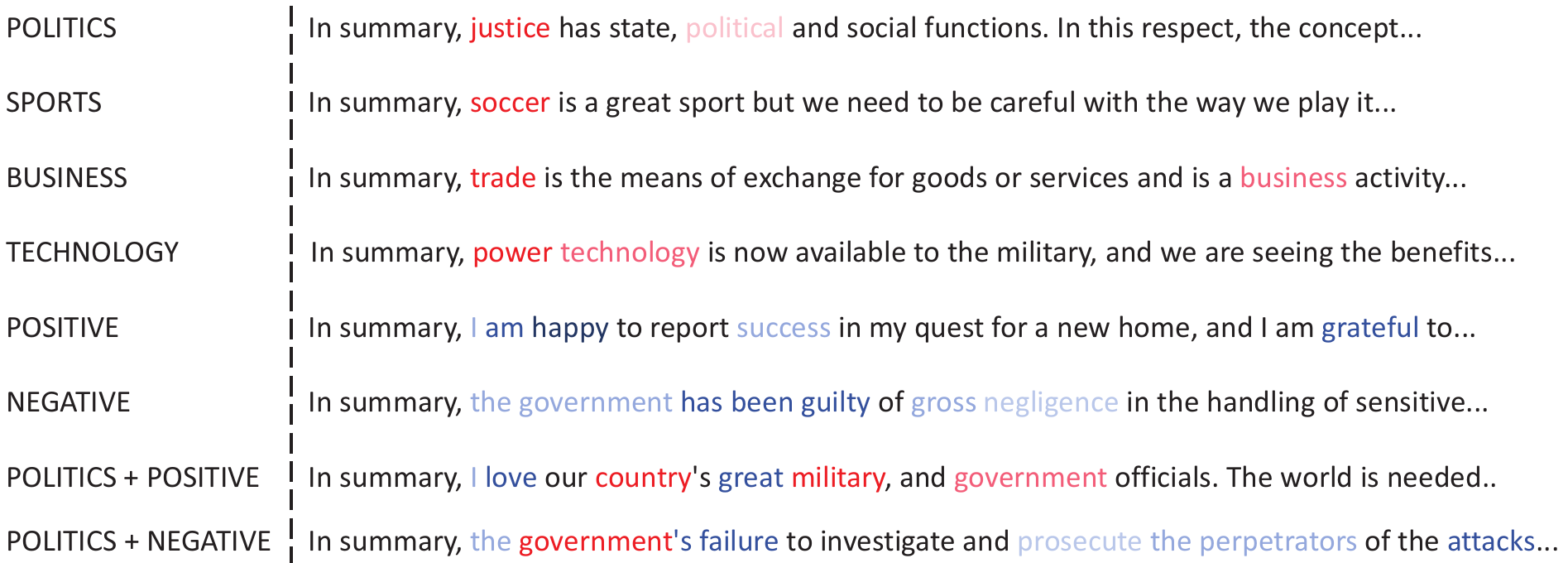}  
\caption{Examples of FreeCtrl's control effects and results. 
The figure employs red to indicate the weights assigned to topic keywords and blue for sentimental keywords, with color saturation corresponding to the weight's intensity.}  
\label{fig_case_study}  
\end{figure*}

To visually demonstrate the control effects, Figure \ref{fig_case_study} displays generation results alongside their respective controlling weights. The figure uses red to denote the weights of topic keywords and blue for the weights of sentimental keywords, with the intensity of each color reflecting the magnitude of the weight.

\section{Inference Speed}
\label{app:inference}

FreeCtrl comprises four main phases: initialization, monitoring, adaptation, and filtering, potentially adding to run-time. Initialization occurs pre-inference, incurring no extra time. Monitoring and adaptation involve evaluating model generation at each timestamp through simple calculations, with negligible added time. Filtering, however, eliminates outputs not meeting certain criteria, leading to wasted generation efforts and additional run-time. We benchmark FreeCtrl's inference time against the SOTA learning-free model, Mix\&Match. We calculate FreeCtrl's average inference time using total run-time in \S\ref{sec:exp} for all outputs (valid and invalid) divided by the number of valid outputs (1225+1400), resulting in an average of 9.8 seconds for FreeCtrl compared to 20 seconds for Mix\&Match. Thus, FreeCtrl not only significantly enhances performance but also substantially reduces inference time.

\section{Scalability}
\label{app:scalability}

We apply FreeCtrl to LLaMA2-7B, aiming to direct the model’s outputs on specified topics including politics, sports, business, and technology. Comparative analysis between the original and FreeCtrl-influenced outputs is detailed in Table \ref{tab_llama}, which demonstrates that FreeCtrl effectively guides LLaMA2’s outputs and significantly enhances attribute relevance scores.

\begin{table}[h]
\centering
\resizebox{0.4\textwidth}{!}{
\begin{tabular}{lcccc}
\hline
\textbf{Method} & \textbf{P} & \textbf{S} & \textbf{B} & \textbf{T}  \\
\hline
Original            & 24.7 & 12.1 & 23.9 &83.9 \\
FreeCtrl        & 81.5 & 83.6 & 79.2 & 98.7 \\
\hline
\end{tabular}
}
\caption{Results of using FreeCtrl on LLaMA2.}
\label{tab_llama}
\end{table}

\end{document}